\documentclass{article} 
\usepackage[preprint]{colm2025_conference}
\usepackage{algpseudocode}
\usepackage{microtype}
\usepackage{hyperref}
\usepackage{url}
\usepackage{multirow}
\usepackage{booktabs}
\usepackage{graphicx}
\usepackage{lineno}
\usepackage{colortbl}
\usepackage{amsmath}
\usepackage{marvosym}
\usepackage{algorithm}
\usepackage{enumitem}
\usepackage{makecell}

\definecolor{darkblue}{rgb}{0, 0, 0.5}
\hypersetup{colorlinks=true, citecolor=darkblue, linkcolor=darkblue, urlcolor=darkblue}
\usepackage[most,skins,theorems]{tcolorbox}

\tcbset{
  aibox/.style={
    width=\linewidth,
    top=8pt,
    bottom=4pt,
    colback=blue!6!white,
    colframe=black,
    colbacktitle=black,
    enhanced,
    center,
    attach boxed title to top left={yshift=-0.1in,xshift=0.15in},
    boxed title style={boxrule=0pt,colframe=white,},
  }
}
\newcolumntype{C}[1]{>{\centering\let\newline\\\arraybackslash\hspace{0pt}}m{#1}}
\newtcolorbox{AIbox}[2][]{aibox,title=#2,#1}

\definecolor{red}{RGB}{238, 68, 51}
\definecolor{blue}{RGB}{70, 177, 225}
\definecolor{yellow}{RGB}{255, 192, 0}
\definecolor{purple}{RGB}{216, 110, 204}
\definecolor{brown}{RGB}{127, 36, 28}
\definecolor{green}{RGB}{71, 172, 20}
\definecolor{orange}{RGB}{194,153,107}

\newcommand{\nothink}[1]{{\color{blue}\textit{#1}}}
\newcommand{\dast}[1]{{\color{yellow}\textit{#1}}}
\newcommand{\efficient}[1]{{\color{brown}\textit{#1}}}
\title{UI-R1: Enhancing Efficient Action Prediction of GUI \\ Agents by Reinforcement Learning}

\author{
Zhengxi Lu$^{1\dag}$, \quad Yuxiang Chai$^{2\dag}$, \quad
Yaxuan Guo$^{1}$, \quad Xi Yin$^{1}$, \\  \,\,\textbf{Liang Liu}$^{1\ddag}$, \quad \textbf{Hao Wang}$^{1}$, \quad \textbf{Han Xiao}$^{2}$,  \quad \textbf{Shuai Ren}$^{1}$, \\ \,\,  \textbf{Guanjing Xiong}$^{1}$,  \quad
\textbf{Hongsheng Li}$^{2}$\textsuperscript{\Letter}
 \\
  \\ \;\textsuperscript{1} vivo AI Lab \quad \textsuperscript{2} MMLab @ CUHK
\\
\small{
$^\dag$~Equal Contribution, \quad $^\ddag$~Project Lead,  \quad \Letter~
    Corresponding Author 
    }
    \\
\small{\{zhengxilu@zju.edu.cn\}}
}

\begin{document}

\ifcolmsubmission
\linenumbers
\fi

\maketitle
\begin{abstract}

The recent DeepSeek-R1 has showcased the emergence of reasoning capabilities in LLMs through reinforcement learning (RL) with rule-based rewards. Despite its success in language models, its application in multimodal domains, particularly in graphic user interface (GUI) agent tasks, remains under-explored. To address this issue, we propose \textbf{UI-R1}, the first framework to explore how rule-based RL can enhance the reasoning capabilities of multimodal large language models (MLLMs) for GUI action prediction tasks. Specifically, UI-R1 introduces a novel rule-based action reward, enabling model optimization via policy-based algorithms such as Group Relative Policy Optimization (GRPO). For efficient training, we curate a small yet high-quality dataset of 136 challenging tasks, encompassing five common action types on mobile devices. Experimental results demonstrate that our proposed \textbf{UI-R1-3B} achieves significant improvements over the base model (i.e. Qwen2.5-VL-3B) on both in-domain (ID) and out-of-domain (OOD) tasks, with average accuracy gains of \textbf{22.1\%} on ScreenSpot, \textbf{6.0\%} on ScreenSpot-Pro, and \textbf{12.7\%} on \textsc{AndroidControl}. Furthermore, UI-R1-3B delivers competitive performance compared to larger models (e.g., OS-Atlas-7B) trained via supervised fine-tuning (SFT) on 76K samples. These results underscore the potential of rule-based reinforcement learning to advance GUI understanding and control, paving the way for future research in this domain. Code website: \url{https://github.com/lll6gg/UI-R1}.
\end{abstract}

\begin{figure}[h]
    \centering

    \includegraphics[width=0.83\textwidth]{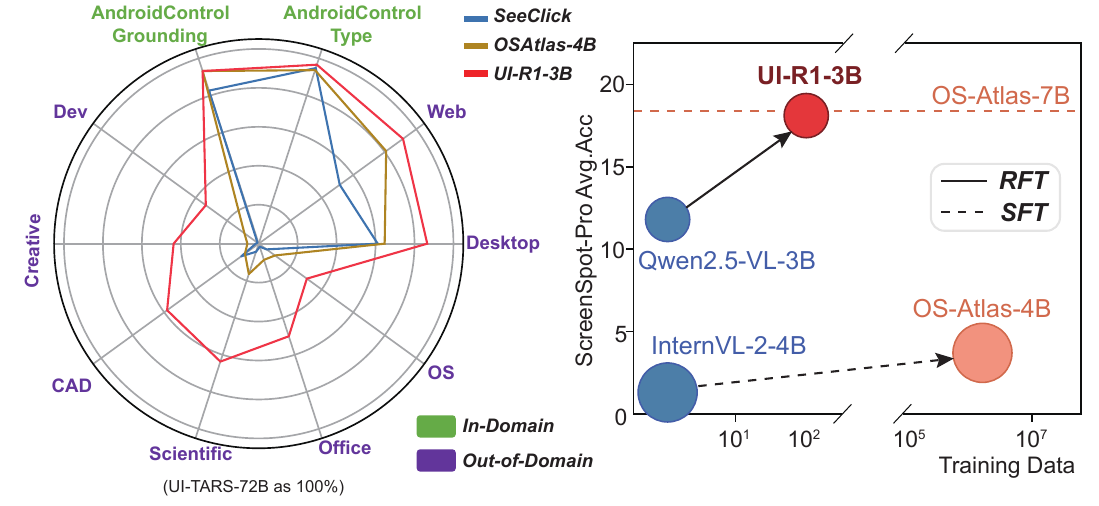}
    \vspace{-1em}
    \caption{\textbf{Left}: Overall performance of UI-R1-3B on both in-domain (i.e., \textsc{AndroidControl}) and out-of-domain (i.e., ScreenSpot-Pro, ScreenSpot desktop and web subsets) tasks; \textbf{Right}: Employing reinforcement fine-tuning (RFT), UI-R1-3B achieves performance comparable to SFT models with significantly fewer data and GPU hours. The circle radius indicates the model size.}
    \label{fig:main_performance}
\end{figure}
\section{Introduction}

Supervised fine-tuning (SFT) has long been the standard training paradigm for large language models (LLMs) and graphic user interface (GUI) agents~\citep{qin2025uitars, wu2024osatlas, hong2024cogagent}. However, SFT relies heavily on large-scale, high-quality labeled datasets, leading to prolonged training times and high computational costs. Furthermore, existing open-source VLM-based GUI agents trained using SFT can be criticized for poor performance in out-of-domain (OOD) scenarios~\citep{lu2024omniparser, chai2024amex}, limiting their effectiveness and applicability in real-world applications.

Rule-based reinforcement learning or reinforcement fine-tuning (RFT) has recently emerged as an efficient and scalable alternative to SFT for the development of LLMs, which efficiently fine-tune the model with merely dozens to thousands of samples to excel at domain-specific tasks. It uses predefined task-specific reward functions, eliminating the need for costly human annotations. Recent works, such as DeepSeek-R1~\citep{guo2025deepseek}, demonstrate the effectiveness of rule-based RL in mathematical problem solving by evaluating the correctness of the solution, while others~\citep{liu2025visual, wang2025visualprm, peng2025lmm, chen2025r1v,huang2025visionr1,zhou2025r1-zero,chen2025visrl} extend the algorithm to multimodal models, achieving notable improvements in vision-related tasks such as image grounding and object detection. By focusing on measurable objectives, rule-based RL enables practical and versatile model optimization across both textual and multimodal domains, offering significant advantages in terms of efficiency, scalability, and reduced reliance on large datasets.

However, existing related studies always target on general vision-related tasks like grounding and detection using Intersection over Union (IoU) metric. In this work, we extend the rule-based RL paradigm to a new application domain by focusing on GUI action prediction tasks driven by low-level instructions. To achieve this, MLLM generates multiple responses (trajectories) that contain the reasoning tokens and the final answers for each input. Then our proposed reward function evaluates each response and updates the model by policy optimization, such as GRPO~\citep{shao2024deepseekmath}. In detail, our action-based reward function contains the action type reward, the action argument reward, along with the commonly used format reward. 
This flexible and effective reward mechanism is well aligned with the objectives of general GUI-related tasks, enhancing model's reasoning capabilities of action prediction by iteratively self-learning.

Regarding data preparation, we follow \citet{muennighoff2025s1} and select just 130+ training mobile samples according to three criterion: difficulty, diversity, and quality, making our method remarkably data-efficient. Experiments demonstrate that UI-R1 achieves significant performance improvements on out-of-domain gounding tasks like ScreenSpot-Pro~\citep{li2024screenspot-pro} and computer scenarios in ScreenSpot~\citep{cheng2024seeclick}, indicating the potential of rule-based RL to tackle complex GUI-related tasks across diverse domains effectively.

In summary, our contributions are as follows. 
\begin{itemize}[leftmargin=5mm]
    \item We propose UI-R1, the \textbf{first} framework which enhances MLLM's reasoning capabilities on GUI action prediction tasks through DeepSeek R1 style reinforcement learning. We believe our exploration can inspire further advancements in the field.
    \item We design a rule-based action reward function that effectively aligns with the objectives of common GUI tasks, facilitating the self-refinement and iterative optimization of the policy model. We also performed some ablation studies to demonstrate its efficiency and rationality.
    \item We utilize the three-stage data selection method and collect only 130+ high-quality training data from the mobile domain. Despite limited data, our proposed UI-R1-3B achieves notable performance gains on out-of-domain benchmarks, such as those from desktop and web platforms, showcasing adaptability and generalization capability in GUI-related tasks.
    \item We additionally develop an optimized version, \textbf{UI-R1-E-3B}, which significantly improves both grounding efficiency and accuracy.
\end{itemize}

\begin{figure}[t]
    \centering
    \includegraphics[width=1.0\textwidth]{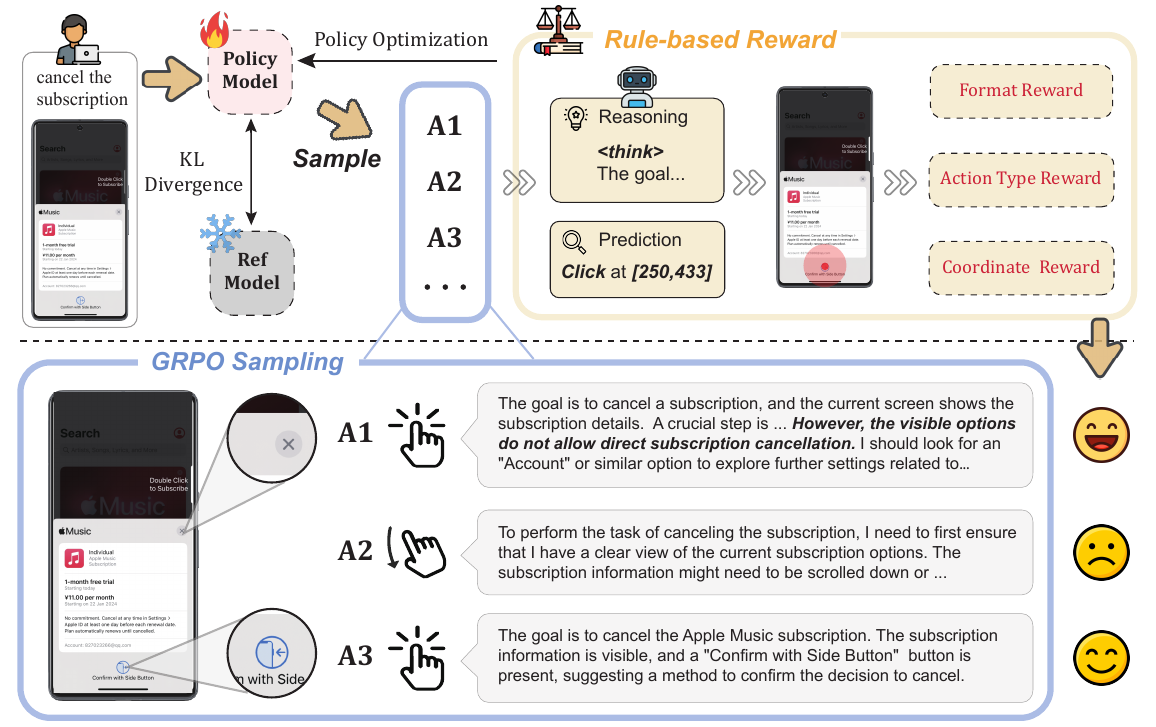}
    \caption{Overview of UI-R1 training framework. Given a GUI screenshot and a text instruction from the user, the policy model (i.e., Qwen2.5-VL-3B) generates multiple action planning responses with reasoning. Our proposed rule-based action reward function is then applied, and the policy model is updated using a policy gradient optimization algorithm.}
    \label{fig:main_method}
\end{figure}

\section{Related Work}  

\subsection{GUI Agents}

Starting with CogAgent~\citep{hong2024cogagent}, researchers have used MLLMs for GUI-related tasks, including device control, task completion, GUI understanding, and more~\citep{liu2025llmsurvey}. One line of work, such as the AppAgent series~\citep{zhang2023appagent, li2024appagentv2} and the Mobile-Agent series~\citep{wang2024mobile, wang2024mobilev2}, integrates commercial generalist models like GPT for planning and prediction tasks. These agents rely heavily on prompt engineering and multi-agent collaboration to execute complex tasks, making them adaptable but dependent on careful manual design for optimal performance. Another branch of research focuses on fine-tuning smaller open-source MLLMs on task-specific GUI datasets~\citep{rawles2023aitw, li2024androidcontrol, chai2024amex, gou2024uground} to create specialist agents. For example, ~\citet{chai2024amex} enhances agents by incorporating additional functionalities of the GUI element in the Android system, while UGround\citep{gou2024uground} develops a special GUI grounding model tailored for precise localization of the GUI element. ~\citet{wu2024osatlas} develops a foundational model for GUI action prediction. Moving beyond task-specific fine-tuning, UI-TARs~\citep{qin2025uitars} introduces a more comprehensive approach by combining GUI-related pretraining with task-wise reasoning fine-tuning, aiming to better align models with the intricacies of GUI interactions. Despite their differences, all of these existing agents share a common reliance on the SFT paradigm. This training approach, while effective, depends heavily on large-scale, high-quality labeled datasets.
 
\subsection{Rule-Based Reinforcement Learning}

Rule-based reinforcement learning has recently emerged as an efficient alternative to traditional training paradigms by leveraging predefined rule-based reward functions to guide model behavior. DeepSeek-R1~\citep{guo2025deepseek} first introduced this approach, using reward functions based on predefined criteria, such as checking whether an LLM's final answer matches the ground truth for math problems. The reward focuses solely on the final results, leaving the reasoning process to be learned by the model itself. ~\citet{zeng2025simplerl} reproduces the algorithm on models with smaller sizes and illustrates its effectiveness on small language models. Subsequent works~\citep{chen2025r1v, shen2025vlmr1, liu2025visual, wang2025visualprm, peng2025lmm, meng2025mm}, extended the paradigm to multimodal models by designing task-specific rewards for visual tasks, including correct class predictions for image classification and IoU metrics for image grounding and detection. These studies demonstrate the adaptability of rule-based RL for both pure-language and multimodal models. By focusing on task-specific objectives without requiring extensive labeled datasets or human feedback, rule-based RL shows strong potential as a scalable and effective training paradigm across diverse tasks.

\paragraph{Efficient Reasoning.}Recent studies have sought to improve reasoning efficiency in large reasoning models (LRMs), addressing challenges such as redundant content (e.g., repeated definitions), over-analysis of simple problems, and shallow exploration of reasoning paths in complex tasks~\citep{qu2025surveyefficient}. To this end, several works~\citep{xiao2025fast,shen2025dast,aggarwal2025l1,li2025bnotthink,wang2025thinkfeet} introduce a \textbf{length reward} within reinforcement learning frameworks, which encourages concise and accurate reasoning by rewarding short, correct answers and penalizing lengthy or incorrect ones. Our work demonstrates that easy tasks like GUI grounding do not need reasoning process and UI-R1-E-3B achieves SOTA performance with fast grounding.

\section{Method}

UI-R1 is a reinforcement learning training paradigm designed to enhance a GUI agent's ability to successfully complete low-level instructional tasks. We define ``low-level instructions'' as directives that guide the agent to perform actions based on a single state (e.g., a GUI screenshot), consistent with the definition in \textsc{AndroidControl}~\citep{li2024androidcontrol}. For example, \textit{``Click the menu icon in the top left corner''} represents a low-level instruction, whereas \textit{``Create an event for 2 PM tomorrow''} is a high-level instruction. The specifics of the training data selection and reward function design are detailed in the following sections. Figure~\ref{fig:main_method} illustrates the main parts of the framework.

\subsection{Preliminary}

Many rule-based RL works~\citep{guo2025deepseek, zeng2025simplerl, liu2025visual} adopt the Group Relative Policy Optimization (GRPO) algorithm~\citep{shao2024deepseekmath} for RL training. GRPO offers an alternative to commonly used Proximal Policy Optimization (PPO)~\citep{schulman2017proximal} by eliminating the need for a critic model. Instead, GRPO directly compares a group of candidate responses to determine their relative quality. 

In GRPO, given a task question, the model generates a set of $N$ potential responses $\{o_1, o_2, \dots, o_N\}$. Each response is evaluated by taking the corresponding actions and computing its reward $\{r_1, r_2, \dots, r_N\}$. Unlike PPO, which relies on a single reward signal and a critic to estimate the value function, GRPO normalizes these rewards to calculate the relative advantage of each response. The relative quality $A_i$ of the $i$-th response is computed as

\begin{equation} \label{eq:grpo}
    A_i = \frac{r_i - Mean(\{r_1, r_2, \dots, r_N\})}{Std(\{r_1, r_2, \dots, r_N\})},
\end{equation}

where $Mean$ and $Std$ represent the mean and standard deviation of the rewards, respectively. This normalization step ensures that responses are compared within the context of the group, allowing GRPO to better capture nuanced differences between candidates.

The policy model is then optimized by maximizing the following KL objective:

\begin{equation}
\begin{split}
    \mathcal{J}_{GRPO}(\theta) = \mathbb{E}_{q \sim P(Q), \{o_i\}_{i=1}^G \sim \pi_{\theta_{old}}(O|q)}
    \Bigg[ \frac{1}{G} \sum_{i=1}^G \frac{1}{|o_i|} \sum_{t=1}^{|o_i|} \Bigg\{ \min \Bigg[
    \frac{\pi_\theta(o_{i,t} | q, o_{i,<t})}{\pi_{\theta_{old}}(o_{i,t} | q, o_{i,<t})} \hat{A}_{i,t},
    \quad \\ \text{clip} \left( \frac{\pi_\theta(o_{i,t} | q, o_{i,<t})}{\pi_{\theta_{old}}(o_{i,t} | q, o_{i,<t})}, 1 - \epsilon, 1 + \epsilon \right)\hat{A}_{i,t}
    \Bigg] - \beta \mathbb{D}_{KL}\left[\pi_{\theta} || \pi_{ref}\right] \Bigg\} \Bigg]
\end{split}
\label{eq:GRPO-obj}
\end{equation}

where $\pi_{\theta}$ and $\pi_{old}$ are the current and old policy, and $\epsilon$ and $\beta$ are hyper-parameters introduced in PPO.


\subsection{Rule-Based Action Rewards}

The rule-based reward function introduced by DeepSeek-R1~\citep{guo2025deepseek} represents a foundational step in rule-based RL by simply evaluating whether model predictions exactly match ground-truth answers. This straightforward approach efficiently aligns models with preference alignment algorithms and provides clear optimization signals. For vision-related tasks, works such as VLM-R1~\citep{shen2025vlmr1} and Visual-RFT~\citep{liu2025visual} extend this idea by designing task-specific rewards. For image grounding tasks, they compute the IoU between the predicted and ground-truth bounding boxes as the reward. Similarly, for image classification tasks, rewards are determined by checking whether the predicted and ground-truth classes match.

In GUI-related tasks, the ability to ground and understand the GUI is a critical requirement for agents. Unlike traditional image grounding tasks, GUI grounding requires agents to identify where specific actions, such as \texttt{click}, should be performed on a given GUI screenshot. To address this unique gap, we propose a reward function tailored for GUI tasks, as defined in Equation~\ref{eq:total_reward}:

\begin{equation}\label{eq:total_reward}
R = R_{\mathcal{T}} + R_{\mathcal{C}} + R_{\mathcal{F}},
\end{equation}

where the predicted action $\mathcal{A}=\{\mathcal{T}, \mathcal{C}\}$ consists of two components: $\mathcal{T}$, which represents the action type (e.g., \texttt{click}, \texttt{scroll}), and $\mathcal{C}$, which represents the \texttt{click} coordinate. $R_\mathcal{F}$ represents the commonly used response format reward.

\paragraph{Action type reward.} 
In our tasks, the action space includes \texttt{Click}, \texttt{Scroll}, \texttt{Back}, \texttt{Open\_App}, and \texttt{Input\_Text}, covering a wide range of common application scenarios in daily life, as inspired by GUIPivot~\citep{wu2025guipivot}. The action type reward, denoted as $\boldsymbol{R_{\mathcal{T}}}$, is computed by comparing the predicted action type $\mathcal{T'}$ with the ground truth action type $\mathcal{T}$. It assigns a reward of 1 if $\mathcal{T'} = \mathcal{T}$ and 0 otherwise, providing a straightforward and effective evaluation mechanism for action type prediction.

\paragraph{Coordinate accuracy reward.}  
Through observation, we find that among all action types, the most common action argument error occurs in the mis-prediction of coordinates for the \texttt{click} action when given a low-level instruction. To address this issue, we specifically design a coordinate accuracy reward. The model is required to output a coordinate $\mathcal{C} = [x, y]$, indicating where the \texttt{click} action should be performed. Given the ground truth bounding box $\mathcal{B} = [x1, y1, x2, y2]$, the coordinate accuracy reward $R_{\mathcal{C}}$ is computed as shown in Equation~\ref{eq:coord_reward}:

\begin{equation} \label{eq:coord_reward}
R_{\mathcal{C}} = 
\begin{cases}
1 & \text{if} \ \text{coord} \ \mathcal{C} \ in \ \text{box} \ \mathcal{B},\\
0 & \text{else}. \\
\end{cases}
\end{equation}

Unlike general visual grounding tasks which compute the IoU between the predicted bounding box and the ground truth box, our approach prioritizes action coordinate prediction over element grounding. This focus is more appropriate for GUI agents and better aligns with human intuition, as the ultimate goal is to ensure correct actions are performed rather than merely locating GUI elements.

\paragraph{Format reward.}

During training, we incorporate the widely-used format reward to guide the model in generating its reasoning process and final answer in a structured format. This decision is based on our simple experiment that agents producing reasoning processes outperform those directly outputting action predictions by approximately 6\% (shown in Appendix~\ref{app:reasoning}). The reasoning process plays a key role in the model's self-learning and iterative improvement during reinforcement fine-tuning, while the reward tied to the final answer drives optimization. The format reward, denoted as $\boldsymbol{R_{\mathcal{F}}}$, ensures that the model's predictions follow the required HTML tag format, specifically using $<$$\boldsymbol{think}$$>$ for the reasoning process and $<$$\boldsymbol{answer}$$>$ for the final answer. This structured output not only enhances clarity, but also ensures consistency in the model's predictions.

\begin{AIbox}{Prompt for Training and Inference}
In this GUI screenshot, I want to perform the command $\boldsymbol{instruction}$. Please provide the action to perform (enumerate in [\texttt{click}, \texttt{open\_app}, \texttt{scroll}, \texttt{navigate\_back}, \texttt{input\_text}]) and the coordinate where the cursor is moved to(integer) if click is performed. Output the thinking process in $<$$\boldsymbol{think}$$>$ $<$$/\boldsymbol{think}$$>$ and final answer in $<$$\boldsymbol{answer}$$>$ $<$$/\boldsymbol{answer}$$>$ tags. The output answer format should be as follows: $<$$\boldsymbol{think}$$>$ ... $<$$/\boldsymbol{think}$$>$ $<$$\boldsymbol{answer}$$>$[{action: enum[\texttt{click}, \texttt{open\_app}, \texttt{scroll}, \texttt{navigate\_back}, \texttt{input\_text}], coordinate: [x, y]}]$<$$/\boldsymbol{answer}$$>$. Please strictly follow the format.
\end{AIbox}
\subsection{Fast Grounding}
Our optimized grounding version, \efficient{UI-R1-E}fficient\efficient{-3B}, is trained in two stages: \dast{DAST} training followed by \nothink{NOTHINK} training, with each stage lasting 4 epochs.
\paragraph{\dast{DAST}} refines rule-based rewards by integrating the deviation between actual response length and the Token Length Budget (TLB) metric, allowing the reward to capture both task difficulty and length properties for effective difficulty-adaptive training~\citep{shen2025dast}.

TLB metric is defined as:

\begin{equation}
L_{budget} = p\cdot L_{\overline{r}} + ( 1-p )\cdot L_{max} ,p = \frac{c}{N}
\end{equation}
where $c$ is the number of correct responses sampled for a given question, and $N$ is the total number of sampled responses. $L_{\overline{r}}$ denotes the average token length of the correct responses, and $L_{\mathrm{max}}$ represents the maximum generation length.

The length reward \dast{$R_{\mathcal{L}}$} is defined as:
\begin{equation}
R_{\mathcal{L}} = 
\begin{cases}
    \max(-0.5 \lambda + 0.5,\ 0.1) & \text{if correct} \\
    \min(0.9 \lambda - 0.1,\ -0.1) & \text{if incorrect}
\end{cases}
\quad \text{where} \quad 
\lambda = \dfrac{L_i - L_{budget}}{L_{budget}}
\label{eq:rl}
\end{equation}

The reward in Equation~\ref{eq:total_reward} is then calibrated by the length reward:

\begin{equation}\label{eq:total_reward_dast}
R = R_{\mathcal{T}} + R_{\mathcal{C}} + R_{\mathcal{F}} + \dast{$R_{\mathcal{L}}$},
\end{equation}

\paragraph{\nothink{NOTHINK}} enables fast and direct grounding by removing the $<$$\boldsymbol{think}$$>$ tags, thereby bypassing explicit reasoning steps during both training and inference~\citep{li2025bnotthink}.
\label{sec:data_selection}

 The corresponding ablation results are presented in Table~\ref{tab:abla_fast_grounding}. Based on these results, we can conclude that 
\begin{center}
\textbf{"Reasoning is not essential for simpler tasks, such as GUI grounding."}
\end{center}
\subsection{Training Data Selection}

Compared to SFT, rule-based RL has demonstrated the capability to achieve comparable or even superior performance on mathematical and vision-related tasks using only a limited number of training samples~\citep{zeng2025simplerl, liu2025visual}. Building on this efficiency and inspired by s1~\citep{muennighoff2025s1}, we implement a three-stage data selection process to refine open-source GUI-related datasets based on three key principles: Quality, Difficulty, and Diversity. The detailed distribution of the dataset can be found in Appendix~\ref{app:data_dist}.

\paragraph{Quality.}
For refining the \texttt{click} action arguments, we use the mobile subset of ScreenSpot~\citep{cheng2024seeclick} as our initial dataset. ScreenSpot offers clean and well-aligned task-element paired annotations, making it ideal for defining and calculating $R_{\mathcal{C}}$. For other actions, we randomly select 1K episodes from \textsc{AndroidControl}~\citep{li2024androidcontrol}, as it shares a similar action space and provides low-level instructions. However, since the element annotations in \textsc{AndroidControl} are unfiltered and misaligned, we exclude \texttt{click} action steps and retain the rest.

\paragraph{Difficulty.}
To identify hard samples, we evaluated Qwen2.5-VL-3B on each task instruction by model performance, where a sample is labeled ``hard'' if the model's output does not match the ground truth. We only keep the ``hard'' samples among all the data collected.

\paragraph{Diversity.}
We ensure diversity by selecting samples with different action types in \textsc{AndroidControl} (e.g., \texttt{Scroll}, \texttt{Back}, \texttt{Open App}, \texttt{Input Text}) and element types in ScreenSpot (e.g. Icon, Text). Rare actions, such as \texttt{Wait} and \texttt{Long Press}, are excluded from \textsc{AndroidControl}. After applying these criteria, we finalize a high-quality mobile training dataset consisting of 136 samples.

\section{Experiments}

\subsection{GUI Grounding Capability}
\label{sec:grounding}

We assess the grounding capability of UI-R1 using two benchmarks: ScreenSpot~\citep{cheng2024seeclick} and ScreenSpot-Pro~\citep{li2024screenspot-pro}. ScreenSpot evaluates GUI grounding capability across mobile, desktop, and web platforms, while ScreenSpot-Pro focuses on high-resolution professional environments, featuring expert-annotated tasks spanning 23 applications, five industries, and three operating systems. We also evaluate the model performance on ScreenSpot-V2~\citep{wu2024osatlas} and the results are in Table~\ref{tab:ss-v2}.

\begin{table}[h]
    \centering
    \resizebox{1.0\textwidth}{!}{
    \renewcommand{\arraystretch}{1.2}
        \begin{tabular}{lcccccccccc}
        \toprule
        \multirow{2}{*}{\textbf{Model}} & \multirow{2}{*}{Method} & Model & Data & \multicolumn{2}{c}{Mobile} & \multicolumn{2}{c}{Web} & \multicolumn{2}{c}{Desktop} & \multirow{2}{*}{Average} \\
        \cmidrule(lr){5-6} \cmidrule(lr){7-8} \cmidrule(lr){9-10}
         & & Size & Size & Icon & Text & Icon & Text & Icon & Text &  \\
        \midrule
        \multicolumn{11}{l}{\textbf{Supervised Fine-tuning}} \\
        \midrule
        CogAgent   & SFT & 18B  & - & 24.0 & 67.0 & 28.6  & 70.4  & 20.0  & 74.2  &\cellcolor{blue!10}47.4 \\
        SeeClick   & SFT & 9.6B & 1M & 52.0 & 78.0 & 32.5  & 55.7  & 30.0  & 72.2  &\cellcolor{blue!10}53.4 \\
        UGround-V1 & SFT & 7B   & 10M & 60.3 & 82.8 & 70.4  & 80.4  & 63.6  & 82.5  & \cellcolor{blue!10}73.3 \\
        Qwen2.5-VL &  SFT & 3B  & 500 & 71.2 & 95.2 & 63.1    &  78.3  &  46.4  &   85.0    &\cellcolor{blue!10}75.7\\
        AGUVIS     & SFT & 7B   & 1M & 78.2 & 88.3 & 70.7  & \underline{88.1}  & \underline{74.8}  & 85.7  & \cellcolor{blue!10}81.8 \\
        \midrule
        \multicolumn{11}{l}{\textbf{Zero Shot / Reinforcement Learning}} \\
        \midrule
        Qwen2-VL   & ZS  & 7B   & 0 & 60.7 & 75.5 & 25.7  & 35.2  & 54.3  & 76.3  &\cellcolor{blue!10}55.3  \\
        Qwen2.5-VL & ZS  & 3B   & 0 & 61.1 & 90.5 & 43.2  & 60.0  & 40.0  & 80.9  &\cellcolor{blue!10}65.0 \\

        UI-R1-3B (Ours) & RFT & 3B & 136 & \textbf{84.7} & \underline{95.6} & \underline{73.3}  & 85.2  & 59.3  & \underline{90.2}  &\cellcolor{blue!10}\underline{83.3}  \\
        \efficient{UI-R1-E-3B} (Ours)& RFT & 3B & 2K & \underline{83.0}  & \textbf{97.1} & \textbf{85.0}  & \textbf{91.7}  & \textbf{77.9}  & \textbf{95.4}  &\cellcolor{blue!10}\textbf{89.2} \\
        \bottomrule
    \end{tabular}
    }
    \caption{Grounding accuracy on ScreenSpot. The optimal and the suboptimal results are \textbf{bolded} and \underline{underlined}, respectively. ZS indicates zero-shot OOD inference and RFT indicates rule-based reinforecement learning.}
    \label{tab:ss}
\end{table}


\begin{table}[h]
    \centering
    \resizebox{0.9\textwidth}{!}{
    \renewcommand{\arraystretch}{1.2}
        \begin{tabular}{lcccccccccccc|c}
        \toprule
        \multirow{2}{*}{\textbf{Model}} & \multirow{2}{*}{GUI specific} & \multirow{2}{*}{Size} & \multicolumn{2}{c}{Mobile} & \multicolumn{2}{c}{Web} & \multicolumn{2}{c}{Desktop} & \multirow{2}{*}{Avg} \\
        \cmidrule(lr){4-5} \cmidrule(lr){6-7} \cmidrule(lr){8-9}
        & & & Icon & Text & Icon & Text & Icon & Text & \\
        \midrule

        SeeClick & Yes & 9.6B & 50.7 & 78.4 & 32.5 & 55.2 & 29.3 & 70.1 & 55.5 \\
        OS-Atlas & Yes & 4B & 59.7 & 87.2 & 63.1 & 85.9 & 46.4 & 72.7 & 71.9 \\
        OS-Atlas & Yes & 7B & 75.8 & 95.2 & 77.3 & \underline{90.6} & 63.6 & 90.7 & 84.1 \\
        UI-TARS & Yes & 2B &79.1 & 95.2 & \underline{78.3} & 87.2 & \underline{68.6} & 90.7 & 84.7 \\
        \midrule
        \multicolumn{8}{l}{\textbf{Qwen2.5-VL Framework}} \\
        \midrule
        Qwen2.5-VL & No  & 3B   & 66.8  & 92.1  & 46.8 & 72.6  & 44.3 & 83.0  & 70.4 \\
        Qwen2.5-VL & No  & 7B   & 80.6  & 95.9  & 70.0  & 87.2  & 59.3  & 89.2 & 82.6 \\
        UI-R1-3B (Ours)& Yes & 3B & \textbf{84.3} & \underline{96.2} & 75.4 &89.2  & 63.6 & \underline{92.3} & \underline{85.4} \\
        \efficient{UI-R1-E-3B} (Ours)& Yes & 3B & \underline{83.9} & \textbf{98.2} & \textbf{83.7 }& \textbf{93.2} & \textbf{75.0} & \textbf{94.8} & \textbf{89.5} \\
        \bottomrule
    \end{tabular}
    }
    \caption{Grounding accuracy on ScreenSpot-V2. The optimal and the suboptimal results are \textbf{bolded} and \underline{underlined}, respectively.}
    \label{tab:ss-v2}
\end{table}

\begin{table}[h]
    \centering
    \resizebox{1,0\textwidth}{!}{%
        \begin{tabular}{lccccccccccccc}
        \toprule
        \multirow{2}{*}{\textbf{Model}} & \multicolumn{2}{c}{\textbf{Development}} & \multicolumn{2}{c}{\textbf{Creative}} & \multicolumn{2}{c}{\textbf{CAD}} & \multicolumn{2}{c}{\textbf{Scientific}} & \multicolumn{2}{c}{\textbf{Office}} & \multicolumn{2}{c}{\textbf{OS}} & \multirow{2}{*}{\textbf{Avg}} \\
        \cmidrule(lr){2-3} \cmidrule(lr){4-5} \cmidrule(lr){6-7} \cmidrule(lr){8-9} \cmidrule(lr){10-11} \cmidrule(lr){12-13} 
         & \textbf{Text} & \textbf{Icon} & \textbf{Text} & \textbf{Icon} & \textbf{Text} & \textbf{Icon} & \textbf{Text} & \textbf{Icon} & \textbf{Text} & \textbf{Icon} & \textbf{Text} & \textbf{Icon} &  \\

        \midrule
        \multicolumn{12}{l}{\textbf{Supervised Fine-tuning}} \\
        \midrule
        SeeClick         & $0.6$  & $0.0$  & $1.0$  & $0.0$  & $2.5$  & $0.0$  & $3.5$  & $0.0$  & $1.1$  & $0.0$  & $2.8$  & $0.0$  & \cellcolor{blue!10}1.1 \\
        OS-Atlas-4B      & $7.1$  & $0.0$  & $3.0$  & $1.4$  & $2.0$  & $0.0$  & $9.0$  & $5.5$  & $5.1$  & $3.8$  & $5.6$  & $0.0$  & \cellcolor{blue!10}3.7 \\
        ShowUI-2B & $16.9$ & $1.4$  & $9.1$  & $0.0$  & $2.5$  & $0.0$  & $13.2$ & $7.3$  & $15.3$ & $7.5$  & $10.3$ & $2.2$  & \cellcolor{blue!10}7.7 \\
        CogAgent-18B & $14.9$ & $0.7$  & $9.6$  & $0.0$  & $7.1$  & $3.1$  & $22.2$ & $1.8$  & $13.0$ & $0.0$  & $5.6$  & $0.0$  & \cellcolor{blue!10}7.7 \\
        Aria-GUI & $16.2$ & $0.0$  & $23.7$ & $2.1$  & $7.6$  & $1.6$  & $27.1$ & $6.4$  & $20.3$ & $1.9$  & 4.7  & $0.0$  & \cellcolor{blue!10}11.3 \\
        Qwen2.5-VL-3B* & 15.6 & 0.7 & 13.1 & 2.1 & 5.6 & 3.1 & 27.8 & 8.1 & 20.3 & 5.7 & 14.0 & 0.0 & \cellcolor{blue!10}10.8 \\
        UGround-7B & 26.6 & $2.1$  & $27.3$ & $2.8$  & 14.2 & $1.6$  & $31.9$ & $2.7$  & $31.6$ & $11.3$ & 17.8 & $0.0$  & \cellcolor{blue!10}16.5 \\
        Claude** & $22.0$ & 3.9  & $25.9$ & 3.4  & \underline{14.5} & $3.7$  & $33.9$ &\underline{15.8}  & $30.1$ & \underline{16.3} & $11.0$ & \underline{4.5}  & \cellcolor{blue!10}17.1 \\
        OS-Atlas-7B & \underline{33.1} & $1.4$  & \underline{28.8} & $2.8$  & $12.2$ & 4.7  & 37.5 & $7.3$  & \underline{33.9} & $5.7$  & \underline{27.1} & \underline{4.5}  & \cellcolor{blue!10}\underline{18.9} \\
        \midrule
        \multicolumn{12}{l}{\textbf{Zero Shot / Reinforcement Fine-tuning}} \\
        \midrule
        Qwen-VL-7B       & $0.0$  & $0.0$    & $0.0$  & $0.0$   & $0.0$  & $0.0$  & $0.7$  & $0.0$   & $0.0$  & $0.0$   & $0.0$  & $0.0$  & \cellcolor{blue!10}0.1 \\
        GPT-4o  & $1.3$  & $0.0$   & $1.0$  & $0.0$   & $2.0$  & $0.0$   & $2.1$  & $0.0$   & $1.1$  & $0.0$   & $0.0$  & $0.0$  & \cellcolor{blue!10}0.8 \\
        Qwen2-VL-7B & $2.6$  & $0.0$   & $1.5$  & $0.0$   & $0.5$  & $0.0$   & $6.3$  & $0.0$  & $3.4$  & $1.9$  & $0.9$  & $0.0$  & \cellcolor{blue!10}1.6 \\
        Qwen2.5-VL-3B & 14.9 & 2.1 & 20.2 & 1.4 & 4.1 & 4.7 & 34.0 & 7.3 & 22.0 & 3.8 & 6.5 & 2.2 & \cellcolor{blue!10}11.8 \\
        \rowcolor{gray!20} UI-R1-3B (Ours) & 22.7 & \underline{4.1} & 27.3 & \underline{3.5} & 11.2 & \underline{6.3} & \underline{42.4} & 11.8 & 32.2 & 11.3 & 13.1 & \underline{4.5} & \cellcolor{blue!10}17.8 \\
        \rowcolor{gray!20} \efficient{UI-R1-E-3B} (Ours) & \textbf{46.1} & \textbf{6.9} & \textbf{41.9} & \textbf{4.2} & \textbf{37.1} & \textbf{12.5} & \textbf{56.9} & \textbf{21.8} & \textbf{65.0} & \textbf{26.4} & \textbf{32.7} & \textbf{10.1} & \cellcolor{blue!10}\textbf{33.5} \\
        \bottomrule
    \end{tabular}
   }
    \caption{Accuracy on ScreenSpot-Pro. The optimal and the suboptimal results are \textbf{bolded} and \underline{underlined}, respectively. * Qwen2.5-VL-3B here is supervised fine-tuned on 500 ScreenSpot-mobile data. ** Claude refers to \textit{Claude-computer-use}.} 
    \vspace{-1em}
    \label{tab:ss-pro}
\end{table}

\paragraph{Setting}
We train the Qwen2.5-VL-3B model on the three-stage selected data (details in Section~\ref{sec:data_selection}) using rule-based RL, naming the resulting model UI-R1-3B. Furthermore, we train the base model using supervised fine-tuning on the entire ScreenSpot mobile set, referring to it as Qwen2.5-VL-3B* in Table~\ref{tab:ss-pro}. For evaluation, an action prediction is considered correct if the predicted \texttt{click} coordinate lies within the ground truth bounding box. Accuracy is computed as the ratio of the correct predictions to the total number of test samples.

\paragraph{Analysis}
Experimental results show that our method significantly improves the GUI grounding capability of the 3B model (\textbf{+20\%} on ScreenSpot and \textbf{+6\%} on ScreenSpot-Pro from Table~\ref{tab:ss} and Table~\ref{tab:ss-pro}), surpassing most 7B models on both benchmarks. Additionally, it also achieves performance comparable to the SOTA 7B models (i.e. AGUVIS~\citep{xu2024aguvis} and OS-Atlas~\citep{wu2024osatlas}), which are trained using supervised fine-tuning on substantially larger labeled grounding datasets.

Qwen2.5-VL-3B (SFT) in Table~\ref{tab:ss} demonstrates that supervised fine-tuning (SFT) with a limited amount of data (e.g., 500 samples) can effectively improve in-domain performance by tailoring the model to specific tasks. However, the comparison between Qwen2.5-VL-3B (ZS) and Qwen2.5-VL-3B (SFT) in Table~\ref{tab:ss-pro} highlights a critical limitation of SFT: its effectiveness significantly diminishes in OOD scenarios. This limitation arises from the dependency of SFT on task-specific labeled data, restricting the model’s ability to adapt to unseen environments. In contrast, our RL approach not only enhances OOD generalization by focusing on task-specific reward optimization, but also achieves with far fewer training samples, offering a scalable and efficient alternative to traditional SFT methods.

\subsection{Action Prediction Capability}
We further evaluate the model's ability to predict single-step actions based on low-level instructions. As described in Section~\ref{sec:data_selection}, we test our model on a selected subset of \textsc{AndroidControl}. The low-level instructions in \textsc{AndroidControl} enrich the ScreenSpot benchmark by introducing a wider range of action types.

\paragraph{Setting}
The accuracy of the action prediction is evaluated by the accuracies of action type and grounding: (1) The action type accuracy evaluates the match rate between the predicted action types (e.g., \texttt{click}, \texttt{scroll}) and ground truth types; (2) The grounding accuracy focuses specifically on the accuracy of \texttt{click} action argument predictions, similar to Section~\ref{sec:grounding}. Since ground truth bounding boxes are not consistently available in the \textsc{AndroidControl} test data, we measure performance by calculating the distance between the predicted and ground truth coordinates. A prediction is considered correct if it falls within 14\% of the screen size from the ground truth, following the evaluation method of UI-TARS~\citep{qin2025uitars}.

\begin{table}[h]
    \centering
    \renewcommand{\arraystretch}{1.2}
    \resizebox{1.0\textwidth}{!}{
    \begin{tabular}{lC{2cm}C{2cm}C{2cm}C{2cm}C{2cm}C{2cm}}
        \toprule
        \textbf{Model} & \textbf{Method} & \textbf{Model size} & \textbf{Data size} & \textbf{Type} & \textbf{Grounding} & \textbf{Average} \\

        \midrule
        \multicolumn{6}{l}{\textbf{Supervised Fine-tuning}} \\
        \midrule
        SeeClick & SFT & 9.6B & 76K & 93.0 & 73.4 &\cellcolor{blue!10}83.2 \\
        InternVL-2 & SFT & 4B &76K & 90.9 & \underline{84.1} &\cellcolor{blue!10}87.5 \\

                GUIPivot-Qwen & SFT & 7B & 76K & \textbf{96.8} & 75.1 & \cellcolor{blue!10}86.0 \\
        OS-Atlas & SFT & 4B & 76K  & 91.9 & 83.8 & \cellcolor{blue!10}87.8 \\
        
        OS-Atlas & SFT & 7B & 76K & 93.6 & \textbf{88.0} & \cellcolor{blue!10}\textbf{90.8} \\

        \midrule
        \multicolumn{6}{l}{\textbf{Zero Shot / Reinforcement Fine-tuning}} \\
        \midrule
        GPT-4o & ZS & -- & 0 & 74.3 & 38.7 & \cellcolor{blue!10}56.5 \\
        OS-Atlas & ZS & 4B & 0 & 64.6 &71.2 &\cellcolor{blue!10}67.9 \\
        OS-Atlas & ZS & 7B & 0 & 73.0 & 73.4 &\cellcolor{blue!10}73.2 \\
        Qwen2.5-VL & ZS  & 3B & 0 & 79.3 & 72.3 & \cellcolor{blue!10}75.8 \\
        UI-R1-3B & RFT & 3B & 136 &\underline{94.3} & 82.6 &\cellcolor{blue!10}\underline{88.5} \\
        \bottomrule
    \end{tabular}
    }
    \caption{Low-level agent capabilities on \textsc{AndroidControl}. The Average column computes the mean of Type and Grounding scores.}
    \vspace{-1em}
    \label{tab:tmr_amr}
\end{table}

\paragraph{Analysis}
As shown in Table~\ref{tab:tmr_amr}, the comparison between UI-R1 and the Qwen2.5-VL (ZS) model highlights the significant benefits of the RL training framework. UI-R1 improves the accuracy of action type prediction by \textbf{15\%} and click element grounding accuracy by \textbf{20\%}, all while using only 136 training data points. Compared with other SFT models, the evaluation results illustrate that UI-R1 not only excels in scenarios with extremely limited training data but also achieves superior type accuracy and grounding performance even than larger models. This underscores the effectiveness of the RL training framework in leveraging small datasets to achieve substantial performance gains, demonstrating its potential as a highly data-efficient and scalable approach for training models in resource-constrained environments.

\subsection{Key Factor Study}

\begin{figure}[t]
    \centering
    \includegraphics[width=1.0\textwidth]{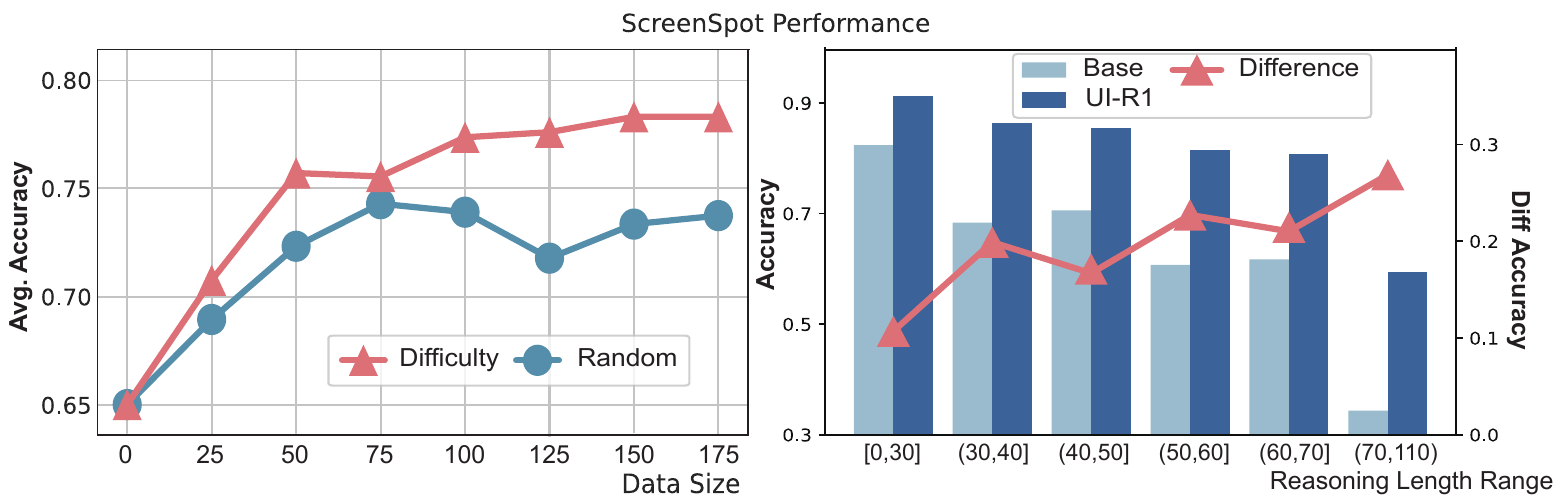}
    \vspace{-2em}
    \caption{\textbf{Left}: Impact of data selection methods and data size; \textbf{Right}: Study of relation between answering accuracy and reasoning length.}
    \vspace{-1em}
    \label{fig:data_size_and_reasoning_lengths}
\end{figure}

\paragraph{Data Size}
    In Figure~\ref{fig:data_size_and_reasoning_lengths} (left), we investigate the relationship between training data size and model performance and compare the two methods of selecting training data from the entire dataset: random selection and \texttt{select by difficulty} (as in Section~\ref{sec:data_selection}). The second method involves selecting the top K tasks with the longest reasoning lengths that the Qwen2.5-VL-3B model fails to solve. We find that model performance improves as the training data size increases, but the trend is gradually saturating. Moreover, our \texttt{select by difficulty} method results in significantly better performance than random selection.

\paragraph{Reasoning Length}
    In Figure~\ref{fig:data_size_and_reasoning_lengths} (right), the result reveals that as the reasoning length of the answer increases, the accuracy tends to decrease, suggesting that the questions are getting harder to answer. With reinforcement learning, UI-R1's reasoning ability is significantly enhanced, leading to more pronounced accuracy improvements, especially on more challenging questions.

\subsection{Ablation Study}
\paragraph{Fast Grounding} The "First \dast{DAST}, then \nothink{NOTHINK}" approach enables the model to learn grounding by gradually transitioning from slower, more thoughtful reasoning to faster, more direct, and accurate responses. The order of these steps is crucial, as disrupting their sequence or removing one of them may significantly impact training efficiency. Moving forward, we aim to explore whether reasoning, as embodied in the thinking phase, is essential for handling high-level planning tasks, or if simpler, direct approaches can yield similar performance in these more complex contexts.
\begin{table}[h]
    \centering
    \resizebox{0.8\textwidth}{!}{
    \renewcommand{\arraystretch}{1.2}
        \begin{tabular}{c|cc|c}
        \toprule
        \textbf{Model} & \textbf{SSV2 Avg\(\uparrow\)} & \textbf{SSPro Avg\(\uparrow\)} & \textbf{Reasoning Length\(\downarrow\)} \\
        \midrule
        \makecell[c]{\dast{DAST} + \nothink{NOTHINK} \\ \scriptsize{(UI-R1-E-3B)}} & \textbf{89.5} & \textbf{33.5} & 0 \\
        \nothink{NOTHINK} + \dast{DAST}              & 87.7          & 31.6          & 0 \\
        \dast{DAST}                        & 87.3          & 30.8          & short \\
        \nothink{NOTHINK}                     & 86.4          & 32.1          & 0 \\
        \textit{THINK} V2                    & 85.7          & 29.8          & long \\
        \textit{THINK} V1 \scriptsize{(UI-R1-3B)}        & 85.4          & 17.8          & long \\
        \bottomrule
        \end{tabular}
    }
    \caption{Abalation of thinking-or-not training methods (\dast{DAST}, \nothink{NOTHINK} or \textit{THINK}) on grounding tasks ScreeSpot V2 (SSV2) and ScreenSpot Pro (SSPro). The total training epoches are all set as 8.}
    \label{tab:abla_fast_grounding}
\end{table}
\paragraph{Reward Function}
The design of the reward function plays a crucial role in enabling the self-learning capabilities of the model. To evaluate this, we first examine the necessity of the two components of the reward function, \texttt{action + coord}. Specifically, the \texttt{action} reward improves action prediction accuracy, while the \texttt{coord} reward enhances the model's ability to ground \texttt{click} elements. Next, we compare this with an alternative reward design, \texttt{action + bbox}, where the coordinate reward $\boldsymbol{R_{\mathcal{C}}}$ is replaced by an IoU-based reward $\boldsymbol{R_{\text{IoU}}}$ in Equation~\ref{eq:total_reward}. In this setup, the IoU metric is calculated between the ground truth bounding box and the predicted box, and $\boldsymbol{R_{\text{IoU}}}$ assigns a value of 1 if $\text{IoU} > 0.5$ and 0 otherwise. 

\begin{figure}[t]
    \centering
    \includegraphics[width=0.96\textwidth]{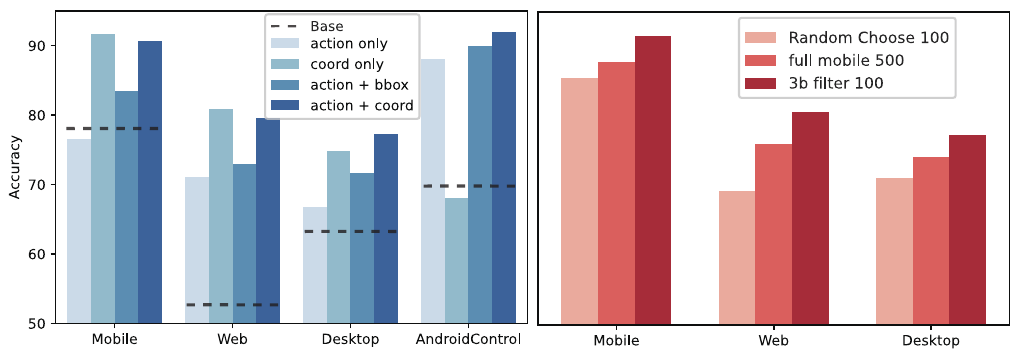}
    \caption{\textbf{Left}: Ablation on reward function; \textbf{Right}: Ablation on data selection method.}
    \label{fig:reward_data_ablation}
\end{figure}

Through ablation studies, as shown in Figure~\ref{fig:reward_data_ablation} (left), we demonstrate the superior effectiveness of $\boldsymbol{R_{\mathcal{C}}}$ over $\boldsymbol{R_{\text{IoU}}}$ for improving \texttt{click} element grounding. However, we also observe that the action reward does not always positively impact grounding tasks. This is likely because a larger action space can introduce ambiguity, making it harder for the model to focus solely on element grounding tasks. These findings highlight the importance of carefully balancing the reward components according to the specific objectives of the task.

\paragraph{Data Selection}
We also examine the impact of different data selection methods, as shown in Figure~\ref{fig:reward_data_ablation} (right). A comparison of three methods across all domains demonstrates that neither random selection nor the use of the entire dataset matches the effectiveness of our three-stage data selection pipeline, indicating that the use of a smaller set of high-quality data can lead to higher performance.


\section{Conclusion}
We propose the UI-R1 framework, which extends rule-based reinforcement learning to GUI action prediction tasks, offering a scalable alternative to traditional Supervised Fine-Tuning (SFT). We designed a novel reward function that evaluates both action type and arguments, enabling efficient learning with reduced task complexity. Using only 130+ training samples from the mobile domain, our proposed UI-R1-3B achieves significant performance improvements and strong generalization to out-of-domain datasets, including desktop and web platforms. The results demonstrate the adaptability, data efficiency, and ability of the rule-based RL to handle specialized tasks effectively.

\bibliography{colm2025_conference}
\bibliographystyle{colm2025_conference}

\newpage
\appendix

\section{Training}
\subsection{Setting}
We configure the hyperparameters as listed in Table~\ref{tab:hyperparams} and train the base model using 8 NVIDIA 4090 GPUs, completing the training process in approximately 8 hours.

\begin{table}[h]
    \centering
    \resizebox{0.6\textwidth}{!}{
    \renewcommand{\arraystretch}{1.2}
    
    \begin{tabular}{c|c}
        \toprule
        \textbf{Hyperparameter} & \textbf{Value} \\ 
        \midrule
        lr & from 9.98e-7 to 0 \\ 
        max\_pixels & 12845056 \\ 
        num\_generations & 8 \\ 
        num\_train\_epochs & 8 \\ 
        max\_prompt\_length & 1024 \\ 
        per\_device\_train\_batch\_size & 1 \\ 
        gradient\_accumulation\_steps & 2 \\ 
        \bottomrule
    \end{tabular}
    }
    \caption{Hyperparameter settings used in the experiments.}
    \label{tab:hyperparams}
\end{table}

\subsection{Dataset Distribution}
\label{app:data_dist}

The distribution of our data selection is listed in Table~\ref{tab:dataset_stats}.

\begin{table}[h!]
    \centering
    \resizebox{1.0\textwidth}{!}{
    \renewcommand{\arraystretch}{1.2}
    
    \begin{tabular}{lccccccc}
        \toprule
        \textbf{Trainng dataset} &Type & \# Click & \# Scroll & \# Input text & \# Back & \# Open app & \# Total \\
        \midrule
        UI-R1 & Mobile & 101 & 5 & 2 & 9 & 19 & 136 \\
        \midrule
        \textbf{Evaluation dataset} \\
        \midrule
        AndroidControl & ID & 5074 & 1211 & 632 & 343 & 608 & 7868\\
        ScreenSpot* & OOD & 770 & 0 & 0 & 0 & 0 & 770 \\
        ScreenSpot-pro & OOD & 1581 & 0 & 0 & 0 &0 & 1581 \\
        \bottomrule
    \end{tabular}
    }
    \caption{Statistics of training and evaluation datasets. * means that we only select subsets Desktop and Web for evaluation.}
    \label{tab:dataset_stats}
\end{table}

\newpage
\subsection{Visualization}
\begin{figure}[h]
    \begin{center}
    \includegraphics[width=1.0 \textwidth]{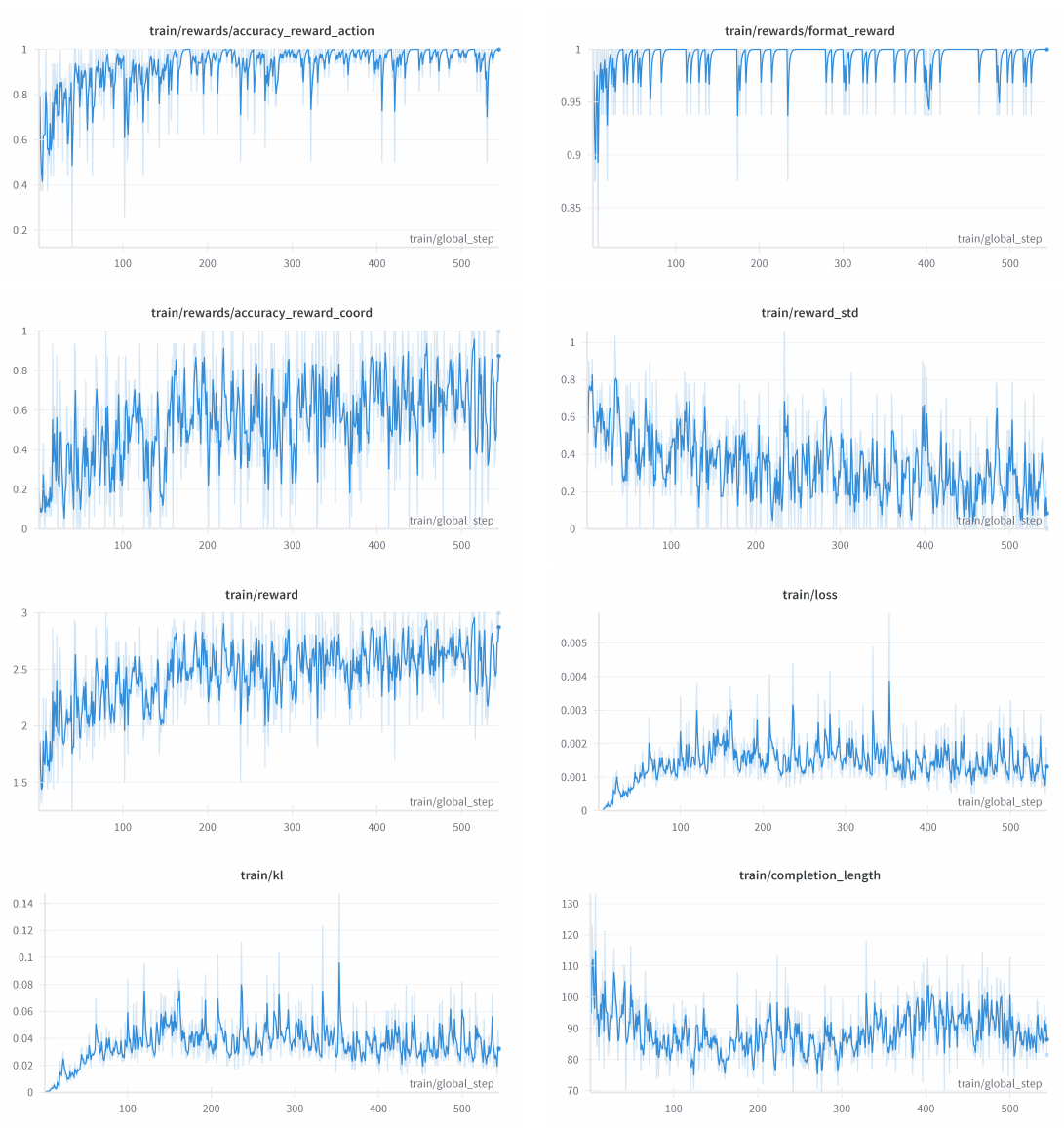}
    \end{center}
    \caption{UI-R1 training process.}
    \label{fig:training_curves}
\end{figure}

Figure~\ref{fig:training_curves} illustrates the progression of various variables throughout the training process.

\newpage
\section{Other Evaluation}
\label{app:more_eval}

\subsection{Reasoning}
\label{app:reasoning}
\begin{figure}[h]
    \begin{center}
    \includegraphics[width=0.7 \textwidth]{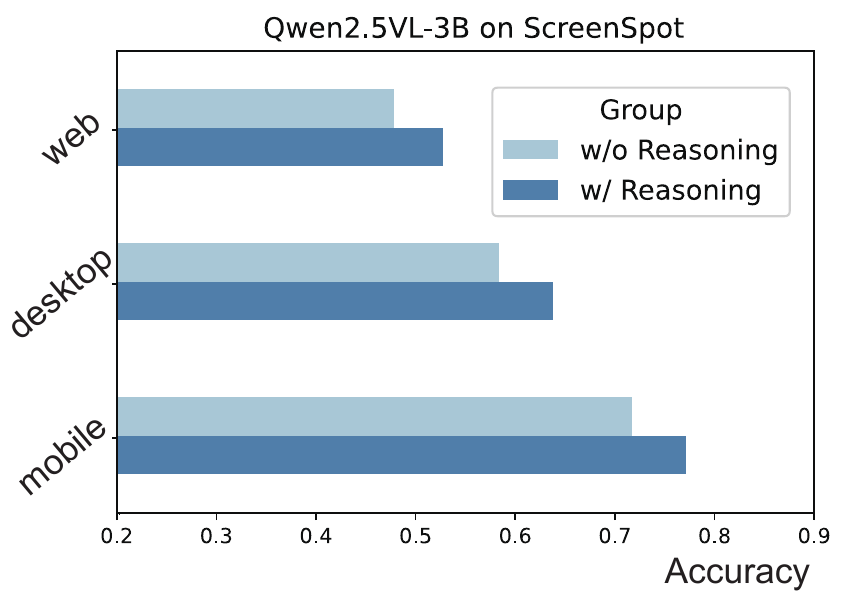}
    \end{center}
    \caption{Qwen2.5-VL-3B's reasoning capability on ScreenSpot.}
    \label{fig:reasoning}
\end{figure}

The reasoning capability of the base model (i.e., Qwen2.5-VL-3B) is central to our approach, as it leverages the model’s self-learning and iterative improvement through reinforcement fine-tuning. To enhance the model's reasoning ability and prediction accuracy, we incorporate reasoning tags into the prompt. To evaluate this, we assess the reasoning performance of Qwen2.5-VL-3B on the ScreenSpot task in Figure~\ref{fig:reasoning}.
\section{Other Ablation}

\subsection{Training epoches}

\begin{figure}[h]
    \begin{center}
    \includegraphics[width=0.9 \textwidth]{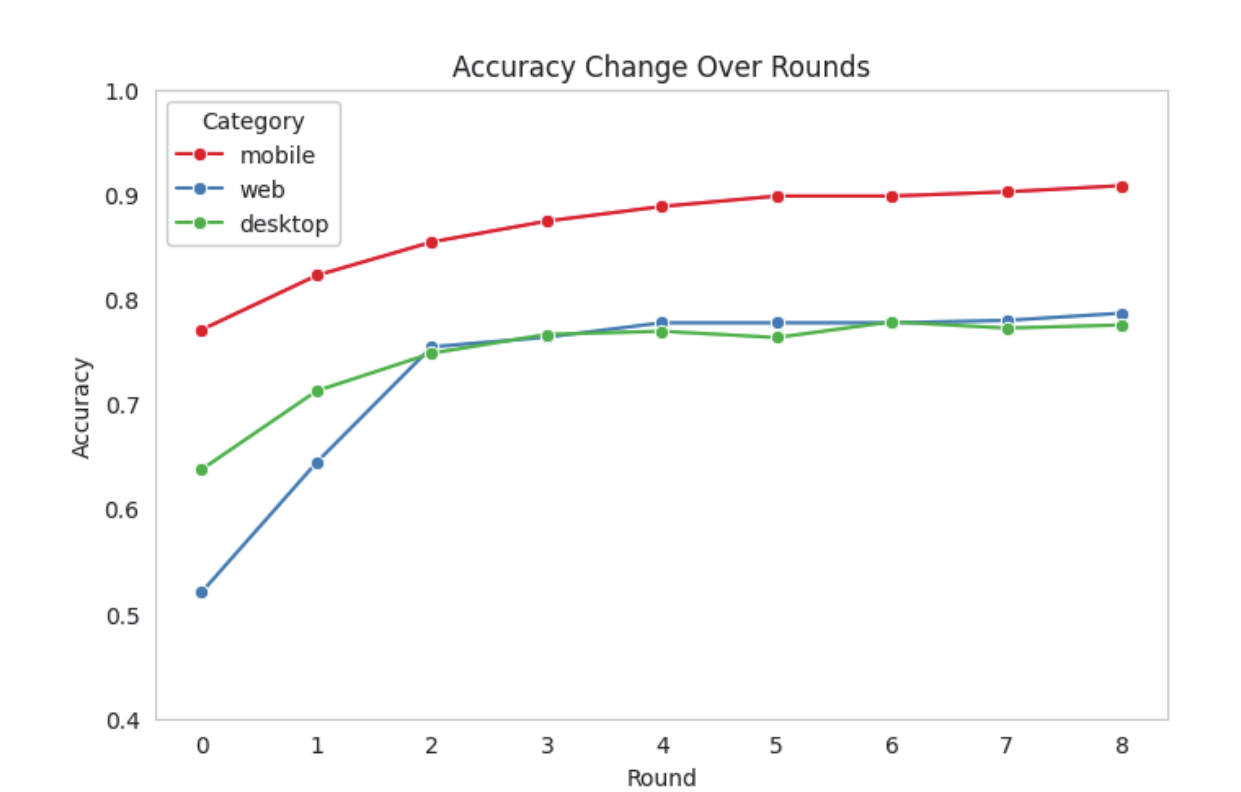}
    \end{center}
    \caption{Accuracy change over rounds.}
    \label{fig:epoch}
\end{figure}

We evaluate the model's performance across different training epochs, as shown in Figure~\ref{fig:epoch}. Based on the results, we finalize the training at 8 epochs.

\subsection{Max Pixels}

While adjusting the parameters, we observe that the maximum pixel setting of the image processor plays a significant role. If the input image exceeds this maximum pixel value, the $\texttt{smart resize}$ function automatically crops and resizes the image while preserving the original aspect ratio. Mobile images are typically smaller than web or desktop images and often have significantly different aspect ratios. To address this, we implement the algorithm to appropriately rescale the predicted coordinates as shown in Algorithm~\ref{alg:scale-coordinates}.

\begin{algorithm}[h]
    \caption{Scale Coordinates Based on Image Resizing}
    \begin{algorithmic}[1]
        \State {\bfseries Input:} 
        \State \ \ \ \ $C = (x, y)$ : coordinate
        \State \ \ \ \ $I$ : input image
        \State \ \ \ \ $max\_pixels$ : maximum pixel constraint
        \State {\bfseries Output:} $(x_{scale}, y_{scale}) \in \mathbb{R}^2$
        \State $(origin\_width, origin\_height) \gets I.size$ 
        \State $(resized\_height, resized\_width) \gets$ \texttt{smart\_resize}$(origin\_height, origin\_width, max\_pixels)$ 
        \textcolor{blue}{\Comment{\texttt{smart\_resize} from QwenVL}}
        \State $x_{scale} \gets origin\_width / resized\_width$  * $x$
        \State $y_{scale} \gets origin\_height / resized\_height$ * $y$
        \State {\bfseries Output:} $(x_{scale}, y_{scale})$
    \end{algorithmic}
    \label{alg:scale-coordinates}
\end{algorithm}
We also investigate the impact of the maximum pixel value on model performance. Setting this value too high can lead to out-of-memory (OOM) errors during training when processing large images. Conversely, setting it too low may negatively affect the accuracy of prediction results. To better understand this trade-off, we experiment with two different maximum pixel values during training and evaluation, as summarized in Table~\ref{tab:max_pixels}.

Based on our analysis, we set the maximum pixel value to 12,845,056 during training, which results in a model with improved performance on out-of-domain tasks. For evaluation, we recommend using a smaller maximum pixel value to conserve memory.

\begin{table}[h]
    \centering
    \renewcommand{\arraystretch}{1.2}
    \begin{tabular}{cc|cccc}
        \toprule
        \multicolumn{2}{c}{max\_pixels} & \multirow{2}{*}{Mobile} & \multirow{2}{*}{Web}& \multirow{2}{*}{Desktop}& \multirow{2}{*}{Avg} \\
        
        Train& Test \\
        \midrule
        3211264  & 3211264  & \textbf{91.2} & 76.1 & 76.6 & 82.2 \\
        3211264  & 12845056 & 90.8 & 76.8 & 76.6 & 82.3 \\
        12845056 & 3211264  & 89.6 & 78.0 & \textbf{77.8} & 82.5 \\
        \rowcolor{gray!20} 12845056 & 12845056 & 90.8 & \textbf{79.6} & 77.2 & \textbf{83.4} \\
        \bottomrule
    \end{tabular}
    \caption{Ablation of max pixels in the training and inference.}
    \label{tab:max_pixels}
\end{table}
\section{Case Study}
\begin{figure}[h]
    \centering
    \includegraphics[width=0.9\textwidth]{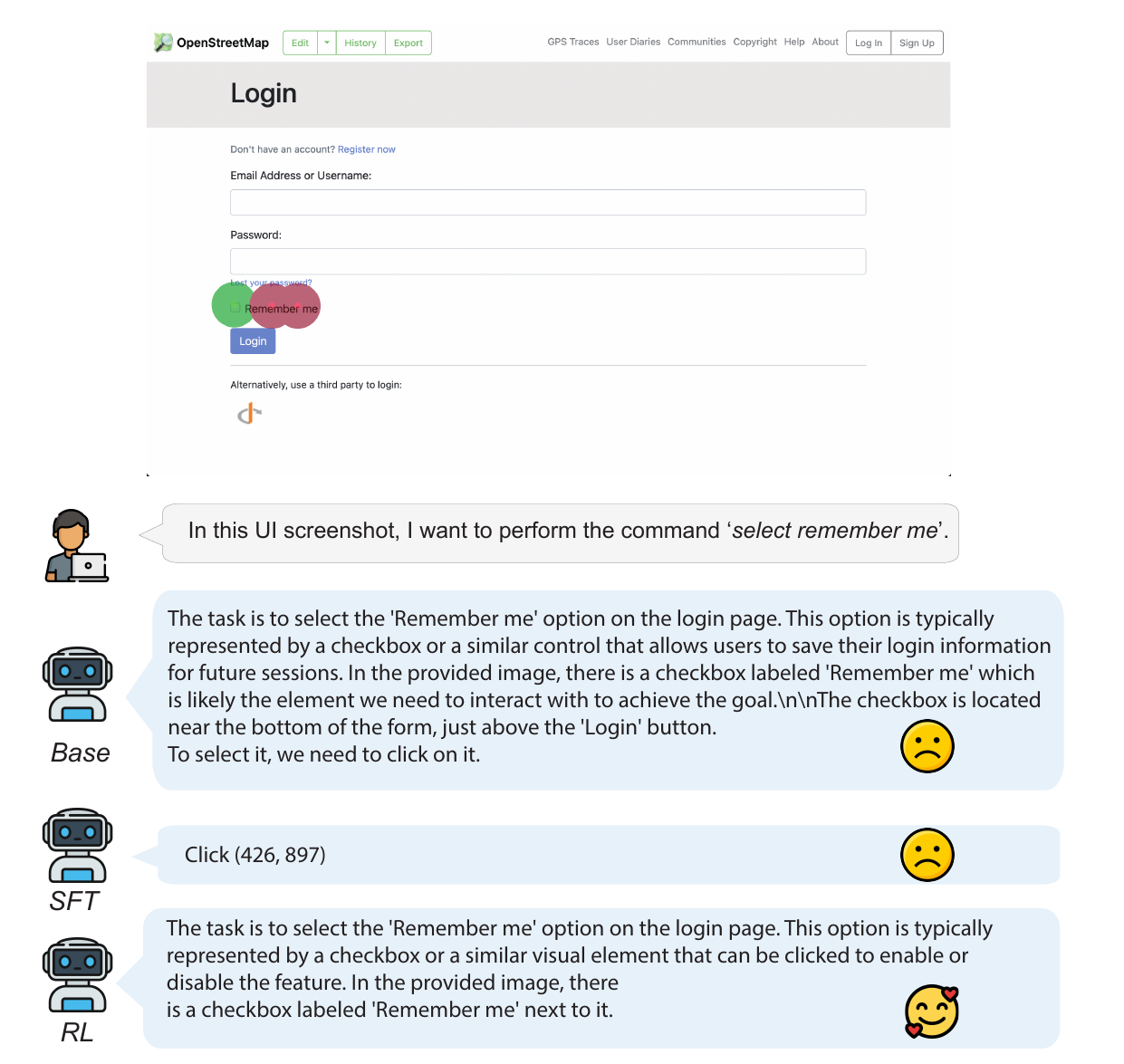}
    \caption{An example of use case.}
    \label{fig:use_case}
\end{figure}

Figure~\ref{fig:use_case} illustrates an example of how UI-R1 trained model can successfully complete the task.

\end{document}